\documentclass[10pt,twocolumn,letterpaper]{article}

\usepackage[pagenumbers]{cvpr} % To force page numbers, e.g. for an arXiv version

\usepackage{graphicx}
\graphicspath{{figure/}}

\usepackage{amsmath}
\usepackage{amsfonts,amssymb}
\usepackage{booktabs}

\usepackage{algorithm}  
\usepackage{algorithmic}  
\usepackage{multirow}
\usepackage[table]{xcolor}
\usepackage{microtype}      % microtypography

\def\pt{\phantom{0}}
\definecolor{grassgreen}{rgb}{0.1,0.8,0.2}

\definecolor{lightgray}{rgb}{0.85, 0.85, 0.85}

\newcommand{\squishlist}{
 \begin{list}{$\bullet$}
  { \setlength{\itemsep}{0pt}
     \setlength{\parsep}{1pt}
     \setlength{\topsep}{1pt}
     \setlength{\partopsep}{0pt}
     \setlength{\leftmargin}{1em}
     \setlength{\labelwidth}{1em}
     \setlength{\labelsep}{0.5em} } }

\newcommand{\squishend}{
  \end{list}  }

\usepackage[pagebackref,breaklinks,colorlinks]{hyperref}

\def\cvprPaperID{2428} % *** Enter the CVPR Paper ID here
\def\confName{CVPR}
\def\confYear{2022}

\begin{document}

\title{Training Vision Transformers with Only 2040 Images}

\author{Yun-Hao Cao, Hao Yu and Jianxin Wu \\
National Key Laboratory for Novel Software Technology \\
Nanjing University, Nanjing, China \\
{\texttt{\{caoyh, yuh\}@lamda.nju.edu.cn, wujx2001@nju.edu.cn}}
}

\maketitle

\begin{abstract}
   Vision Transformers (ViTs) is emerging as an alternative to convolutional neural networks (CNNs) for visual recognition. They achieve competitive results with CNNs but the lack of the typical convolutional inductive bias makes them more data-hungry than common CNNs. They are often pretrained on JFT-300M or at least ImageNet and few works study training ViTs with limited data. In this paper, we investigate how to train ViTs with limited data (e.g., 2040 images). We give theoretical analyses that our method (based on parametric instance discrimination) is superior to other methods in that it can capture both feature alignment and instance similarities. We achieve state-of-the-art results when training from scratch on 7 small datasets under various ViT backbones. We also investigate the transferring ability of small datasets and find that representations learned from small datasets can even improve large-scale ImageNet training.
\end{abstract}

\section{Introduction}
\label{sec:intro}

Transformers~\cite{transformer:Vaswani:NIPS17} have recently emerged as an alternative to convolutional neural networks (CNNs) for visual recognition~\cite{vit:dosovitskiy:ICLR21, DeiT:Touvron:ICML2021, t2t:yuan:arxiv2021}. The vision transformer (ViT) introduced by Dosovitskiy \etal~\cite{vit:dosovitskiy:ICLR21} is an architecture directly inherited from natural language processing~\cite{bert:devlin:NAACL19}, but applied to image classification with raw image patches as input. ViT and variants achieve competitive results with CNNs but require significantly more training data. For instance, ViT performs worse than ResNets~\cite{resnet:he:CVPR16} with similar capacity when trained on ImageNet~\cite{ILSVRC2012:russakovsky:IJCV15} (1.28 million images). One possible reason may be that ViT lacks certain desirable properties inherently built into the CNN architecture that make CNNs uniquely suited to solve vision tasks, e.g., locality, the translation invariance and the hierarchical structure~\cite{cvt:wu:arxiv2021}. As a result, ViTs need a lot of data for training, usually more data-hungry than CNNs. 

In order to alleviate this problem, a lot of works try to introduce convolutions to ViTs~\cite{cvt:wu:arxiv2021, pvt:wang:arxiv2021, swin:liu:arxiv2021, t2t:yuan:arxiv2021}. These architectures enjoy the advantages of both paradigms, with attention layers modeling long-range dependencies while convolutions emphasizing the local properties of images. Empirical results show that these ViTs trained on ImageNet outperform similar-size ResNets on this dataset. However, ImageNet is still a large-scale dataset and it is still not clear what is the behavior of these networks when trained on small datasets (e.g., 2040 images). As will be further analysed in Sec.~\ref{sec:why}, we cannot always rely on such large-scale datasets from the perspective of data, computing and flexibility.

In this paper, we investigate how to train ViTs \emph{from scratch} with limited data. We first perform self-supervised pretraining and then supervised fine-tuning on the same target dataset, as done in~\cite{S3L:cao:arxiv2021}. We focus on the self-supervised pretraining stage and our method is based on parametric instance discrimination~\cite{exemplar:alexey:nips14}. We theoretically analyze that parametric instance discrimination can not only capture feature alignment between positive pairs but also find potential similarities between instances thanks to the final learnable fully connected layer $W$. Experimental results further verify our analyses and our method achieves better performance than other non-parametric contrastive methods~\cite{simclr:hinton:ICML20, mocov2:xinlei:arxiv2020, mocov3:chen:ICCV21, dino:caron:iccv2021}. It is known that instance discrimination suffers from high GPU computation, high memory overload and slow convergence for high-dimensional $W$ on large-scale datasets. Since in this paper we focus on small datasets, we do not need complicated strategies for large-scale datasets as in~\cite{ParametricInstance:cao:arxiv2020, onemillion:liu:AAAI21}. Instead, we adopt small resolution~\cite{S3L:cao:arxiv2021}, multi-crop~\cite{swav:caron:NIPS20} and CutMix~\cite{cutmix:yun:ICCV19} for the small data setup and we also analyze them from both the theoretical and empirical perspectives.

We name our method as \textbf{I}nstance \textbf{D}iscrimination with \textbf{M}ulti-crop and Cut\textbf{M}ix (IDMM) and achieve state-of-the-art results on 7 small datasets when training from scratch under various ViT backbones. For instance, we achieve 96.7\% accuracy when training from scratch on flowers~\cite{flowers} (2040 images), which shows that training ViTs with small data is surprisingly viable. Moreover, we first analyze the transferring ability of small datasets. We find that ViTs also have good transferring ability even when pretrained on small datasets and can even facilitate training on large-scale datasets, e.g., ImageNet. Liu \etal~\cite{ViT-small-data:liu:arxiv2021} also investigate training ViTs with small-size datasets but they focus on the fine-tuning stage while we focus on the pretraining stage. More importantly, we achieve much better results than~\cite{ViT-small-data:liu:arxiv2021}, where the best reported accuracy on flowers was 56.3\%. 

In summary, our contributions are:

\squishlist
    \item We propose IDMM for self-supervised ViT training and achieve state-of-the-art results even when training from scratch for various ViT backbones on 7 small datasets.
    \item We give theoretical analyses on why we should prefer parametric instance discrimination when dealing with small data from the loss perspective. Moreover, we show how strategies like CutMix alleviate the infrequent updating problem from the gradient perspective. 
    \item We empirically show that the projection MLP head is essential for non-parametric contrastive methods (e.g., SimCLR~\cite{simclr:hinton:ICML20}) but not for parametric instance discrimination, thanks to the final learnable $W$ in instance discrimination.
    \item We analyze the transferring ability of small datasets and find that ViTs also have good transferring ability even when pretrained on small datasets. 
\squishend

\section{Related Works}

\noindent\textbf{Self-supervised learning.} Self-supervised learning (SSL) has emerged as a powerful method to learn visual representations without labels. Many recent works follow the contrastive learning paradigm~\cite{InfoNCE:arxiv2018}, which is also known as non-parametric instance discrimination~\cite{memorybank:wu:CVPR18}. For instance, SimCLR~\cite{simclr:hinton:ICML20} and MoCo~\cite{moco:kaiming:CVPR20} trained networks to identify a pair of views originating from the same image when contrasted with many views from other images. Unlike the two-branch structure in contrastive methods, some approaches~\cite{exemplar:alexey:nips14, ParametricInstance:cao:arxiv2020, onemillion:liu:AAAI21} employ a parametric, one-branch structure for instance discrimination. Exemplar-CNN~\cite{exemplar:alexey:nips14} learned to discriminate between a set of surrogate classes, where each class represents different transformed patches of a single image. \cite{ParametricInstance:cao:arxiv2020} and \cite{onemillion:liu:AAAI21} proposed different methods to alleviate the infrequent instance visiting problem or reduce the GPU memory consumption for large-scale datasets, but rely on complicated engineering techniques for CNNs and lack theoretical analyses. In this paper, we not only apply parametric instance discrimination to ViTs, but also focus on small datasets. In addition, we give theoretical analyses of why should we prefer parametric method, at least for small datasets.

Recently, there have also been self-supervised methods designed for ViTs. \cite{mocov3:chen:ICCV21} found that instability is a major issue that impacts self-supervised ViT training and proposed a simple contrastive baseline MoCov3. DINO~\cite{dino:caron:iccv2021} designed a simple self-supervised approach that can be interpreted as a form of knowledge distillation with no labels. However, they focused on large-scale datasets while we focus on small data. Our method is more stable for various networks and more effective for small data. 

\noindent\textbf{Vision Transformers.} Vision Transformer (ViT)~\cite{vit:dosovitskiy:ICLR21} treated an image as patches/tokens and employed a pure transformer structure. With \emph{sufficient} training data,  ViT outperforms CNNs on various image classification benchmarks, and many ViT variants have been proposed since then. Touvron \etal \cite{DeiT:Touvron:ICML2021} introduced a teacher-student distillation token strategy into ViT, namely DeiT. Beyond classification, Transformer has been adopted in diverse vision tasks, including detection~\cite{detr:carion:ECCV2020}, segmentation~\cite{VisTR:wang:CVPR2021}, etc. Many ViT variants were proposed in recent months. Swin Transformer~\cite{swin:liu:arxiv2021} applied the shifted window approach to compute self-attention matrix. Wang \etal proposed PVT-based model (PVTv1 \& v2)~\cite{pvt:wang:arxiv2021, pvtv2:wang:arxiv2021}, which built a progressive shrinking pyramid and a spatial-reduction attention layer to generate multi-resolution feature maps. T2T-ViT~\cite{t2t:yuan:arxiv2021} introduced a tokens-to-token (T2T) module to aggregate neighboring tokens into one recursively. However, ViTs are known to be data-hungry~\cite{ViT-small-data:liu:arxiv2021} and how to train ViTs with limited data is an important but not fully investigated question. \cite{ViT-small-data:liu:arxiv2021} proposed a self-supervised task for ViTs, which can extract additional information from images and make training much more robust when training data are scarce. In contrast, we focus on the self-supervised pretraining stage while~\cite{ViT-small-data:liu:arxiv2021} focuses on the supervised fine-tuning stage. Moreover, we achieve much higher accuracy when training from scratch and we investigate the transferring ability when training on small datasets.

\section{Method}

We first explain why we use parametric instance discrimination (Sec.~\ref{sec:analysis}), then analyze how our strategies help weight updating (Sec.~\ref{sec:gradient}), and describe the complete method.

\subsection{Analyses on instance discrimination}
\label{sec:analysis}

\begin{figure}
	\centering
	\includegraphics[width=0.7\columnwidth]{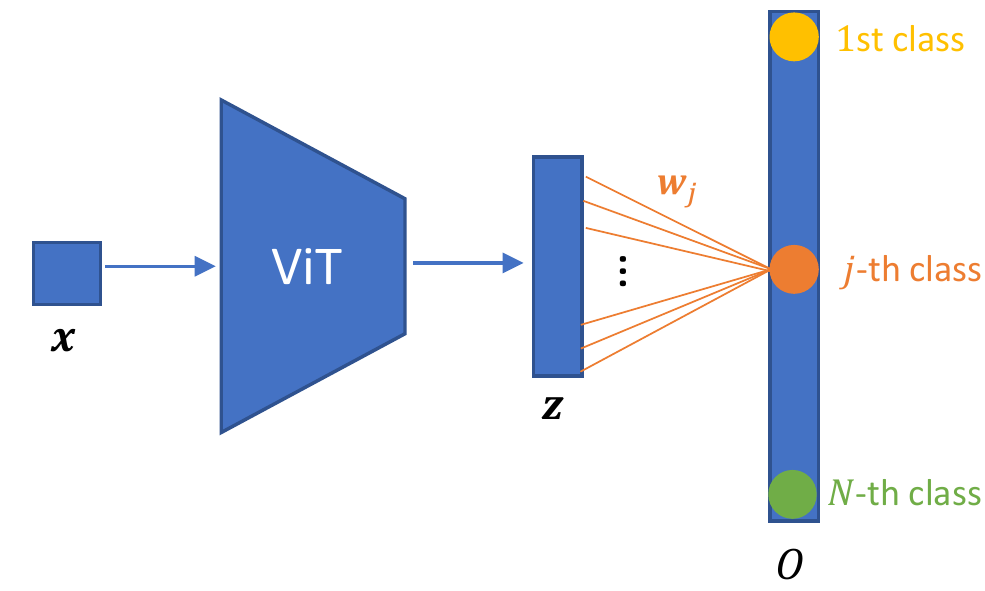}
	\vspace{-6pt}
	\caption{Illustration of parametric instance discrimination.}
	\label{fig:insdis}
\end{figure}

As shown in Figure~\ref{fig:insdis}, an input image $\boldsymbol{x}_i$ ($i=1,\cdots,N$) is sent to a network $f(\cdot)$ and get output representation $\mathbf{z}_i=f(\boldsymbol{x}_i)\in{\mathbb{R}^d}$, where $N$ denotes the total number of instances. Then, a fully connected (fc) layer $W$ is used for classification and the number of classes equals the total number of training images $N$ for parametric instance discrimination. We denote $\mathbf{w}_j\in{\mathbb{R}^{d}}$ as the weights for the $j$-th class and $W=[\mathbf{w}_1 | \dots | \mathbf{w}_N]\in{\mathbb{R}^{d\times{N}}}$ contains the weights for all $n$ classes. Hence we have $O^{(i)}=W^T\mathbf{z}_i$, where the output for the $j$-th class $O^{(i)}_j=\mathbf{w}^T_j{\mathbf{z}_i}$. Finally, $O^{(i)}$ is sent to a softmax layer to get a valid probability distribution $P^{(i)}$.

For instance discrimination, the loss function is:
\begin{align}
    L_{\text{InsDis}} &= -\sum_{i=1}^N\sum_{c=1}^N{y^{(i)}_c\log{P^{(i)}_c}} = -\sum_{i=1}^N \log{P^{(i)}_i} \\
    &= -\sum_{i=1}^N \log\frac{\exp(\mathbf{w}_i^T\mathbf{z}_i)}{\sum_{j=1}^N\exp(\mathbf{w}_j^T\mathbf{z}_i)} \\
    &= -\sum_{i=1}^N \mathbf{w}_i^T\mathbf{z}_i+\sum_{i=1}^N\log\sum_{j=1}^Ne^{\mathbf{w}_j^T\mathbf{z}_i} \,,\label{eq:loss_insdis}
\end{align}
where the superscript $i$ sums over instances while the subscript $c$ sums over classes. For instance discrimination, the class label corresponds to the instance ID: $y_c^{(i)}=1 \text{ iff } c=i$.

Now we move on to the contrastive learning (CL) loss. There are typically 2 views (i.e., positive pairs) for each input $\boldsymbol{x}_i$ and we call them $\boldsymbol{x}_{iA}$, $\boldsymbol{x}_{iB}$ (corresponding representations are $\mathbf{z}_{iA}$, $\mathbf{z}_{iB}$). The contrastive loss can be represented as follows (we omit hyper-parameter $\tau$ for simplicity):
\begin{align*}
    L_{CL} =& -\sum_{i=1}^N \log\frac{e^{\mathbf{z}_{iA}^T\mathbf{z}_{iB}}}{e^{\mathbf{z}_{iA}^T\mathbf{z}_{iB}}+\sum_{i}e^{\mathbf{z}_{iA}^T\mathbf{z}_{i}^-}}\\
    =&-\sum_{i=1}^N{\mathbf{z}_{iA}^T\mathbf{z}_{iB}}+
    \sum_{i=1}^N\log\left(e^{\mathbf{z}_{iA}^T\mathbf{z}_{iB}}+\sum e^{\mathbf{z}_{iA}^T\mathbf{z}_{i}^-} \right)\,,
\end{align*}
where $\mathbf{z}_i^-$ enumerates all negative pairs for $\mathbf{z}_i$, i.e., $\mathbf{z}_{jA}$ and $\mathbf{z}_{jB}$ for all $j\neq{i}$. Consider the loss term for the $i$-th instance:
\begin{equation}
\label{eq:CL_i}
    L^{(i)}_{CL}=\underbrace{-\mathbf{z}_{iA}^T\mathbf{z}_{iB}}_{\text{alignment}}+\underbrace{\log\left(e^{\mathbf{z}_{iA}^T\mathbf{z}_{iB}}+\sum e^{\mathbf{z}_{iA}^T\mathbf{z}_{i}^-} \right)}_{\text{uniformity}}
\end{equation}

If we set $\mathbf{w}_i=\mathbf{z}_i$ in instance discrimination, then from Eq.~\ref{eq:loss_insdis} we have (also consider the $i$-th term):
\begin{equation}
\label{eq:InsDis_i}
    L^{(i)}_{\text{InsDis}}=\underbrace{-\mathbf{z}_i^T\mathbf{z}_{i}}_{\text{alignment}}+\underbrace{\log\left(e^{\mathbf{z}_i^T\mathbf{z}_i}+\sum\nolimits_{j\neq{i}}e^{\mathbf{z}_i^T\mathbf{z}_{j}}\right)}_{\text{uniformity}}
\end{equation}

Now it is clear that Eqs.~\eqref{eq:InsDis_i} and~\eqref{eq:CL_i} are almost identical, except that there are two views in Eq.~\eqref{eq:CL_i} ($\mathbf{z}_{iA}$ and $\mathbf{z}_{iB}$ vs. $\mathbf{z}_i$). Both have two terms: the alignment term encouraging more aligned positive features and the uniformity term encouraging the features to be roughly uniformly distributed on the unit hypersphere, as noted in~\cite{hypersphere:wang:ICML20}. Hence, we conclude that instance discrimination is approximately equivalent to the contrastive loss when we set $\mathbf{w}_j=\mathbf{z}_{j}, \forall\,{j}$. Our analyses also give a theoretical interpretation of the contrastive prior used in~\cite{onemillion:liu:AAAI21}, which initializes $W$ in a contrastive way to accelerate convergence for high-dimensional $W$.

Moreover, we can also use multiple views in instance discrimination. If we set $\mathbf{w}_j=\mathbf{z}_{jA}$, we have (see the appendix for detailed derivation):
\begin{align}
    L^{(i)}_{\text{InsDis}}=&\underbrace{-\mathbf{z}_{iA}^T\mathbf{z}_{iB}}_{\text{alignment}}
    +\underbrace{\log\left(e^{\mathbf{z}_{iA}^T\mathbf{z}_{iB}}+\sum\nolimits_{j\neq{i}}e^{\mathbf{z}_{iB}^T\mathbf{z}_{jA}}\right)}_{\text{uniformity}}\\
    &\underbrace{-\mathbf{z}_{iA}^T\mathbf{z}_{iA}}_{\text{alignment}}+\underbrace{\log\left(e^{\mathbf{z}_{iA}^T\mathbf{z}_{iA}}+\sum\nolimits_{j\neq{i}}e^{\mathbf{z}_{iA}^T\mathbf{z}_{jA}}\right)}_{\text{uniformity}}
\end{align}

In other words, the contrastive loss is a special case of instance discrimination, with each $\mathbf{w}_i$ set to the representation of $\mathbf{x}_i$ in the current batch (i.e., non-parametric instance discrimination). In contrast, the learnable fc $W$ in instance discrimination has at least two advantages:

(i) Separate representation learning from learning specific properties of the loss. As known in many contrastive learning methods (e.g., SimCLR~\cite{simclr:hinton:ICML20}), using extra projection head (MLPs) after representation is essential to learn good representations. However, we find that this projection head is \emph{not} necessary for instance discrimination, thanks to the learnable weights $W$ of this fc, as will be shown in Section~\ref{sec:ablation}.

(ii) Find potential similarities between instances (classes). Now we consider DeepClustering~\cite{deepclustering:caron:ECCV18}, whose clustering loss can be reformulated as follows using our notation:
\begin{equation}
\label{eq:deep-clustering}
    L_{\text{DC}} = -\sum_{i=1}^N\sum_{k=1}^K y_k^{(i)}\log{P_k^{(i)}}\,,
\end{equation}
where $K$ denotes the number of clusters, $y^{(i)}_k$ indicates whether the $i$-th instance belongs to the $k$-th cluster, and $P^{(i)}_k$ denotes the probability that the $i$-th instance belongs to the $k$-th cluster. Let $C_k$ denotes the index of instances in cluster $k$, then if we set all $\{\mathbf{w}_j|j\in{C_k}\}$ to the same, i.e., $\mathbf{w}_j=\Tilde{\mathbf{w}}_k$ for all $j\in C_k$, we have:
\begin{align}
    L_{\text{InsDis}} =& -\sum_{i=1}^N \log P_i^{(i)} = -\sum_{k=1}^K \sum_{j\in{C_k}} \log{P_j^{(j)}} \\
    =&-\sum_{k=1}^K\sum_{j\in{C_K}} \log\sigma(\mathbf{w}_j^T\mathbf{z}_j)\\
    =&-\sum_{k=1}^K\sum_{j\in{C_K}}\log\sigma(\Tilde{\mathbf{w}}_k^T\mathbf{z}_j)\,,
 \end{align}
$\sigma(\cdot)$ is the softmax function. Similarly, Eq.~\eqref{eq:deep-clustering} becomes
\begin{equation}
    L_{\text{DC}} = -\sum_{k=1}^K \sum_{j\in{C_k}} \log{P_k^{(j)}} 
    =-\sum_{k=1}^K\sum_{j\in{C_K}}\log\sigma(\Tilde{\mathbf{w}}_k^T\mathbf{z}_j)\,.
\end{equation}

\begin{figure*}[!htbp]
    \centering
	\vspace{-12pt}
    \subfloat[random init.]{
        \label{fig:tSNE-random}
        \includegraphics[width=0.275\linewidth]{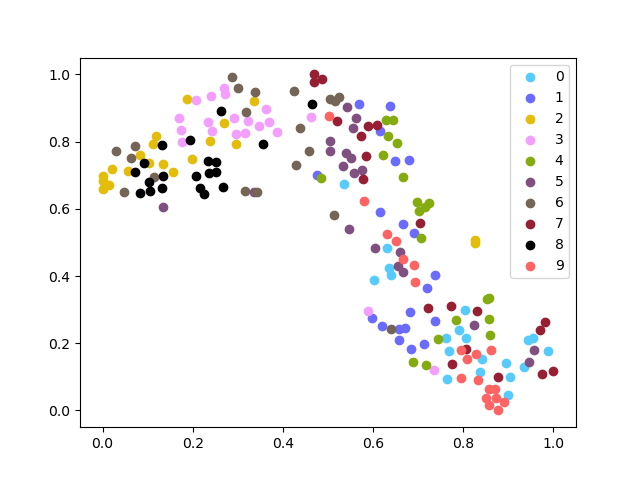}
	}
    \subfloat[SimCLR]{
        \label{fig:tSNE-simclr}
        \includegraphics[width=0.275\linewidth]{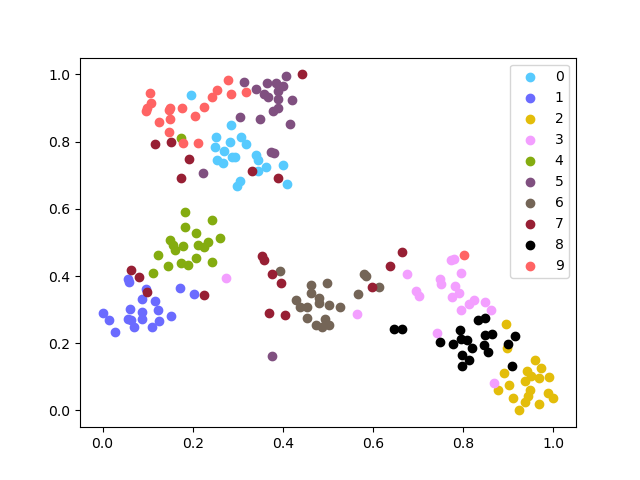}
	}
    \subfloat[IDMM (ours)]{
        \label{fig:tSNE-InsDis}
        \includegraphics[width=0.275\linewidth]{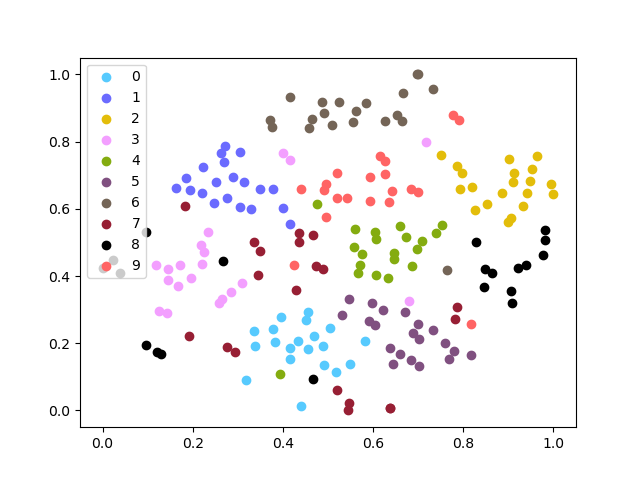}
    }
    \\
	\vspace{-2pt}
	\subfloat[random init.+FT]{
        \label{fig:tSNE-random-FT}
        \includegraphics[width=0.275\linewidth]{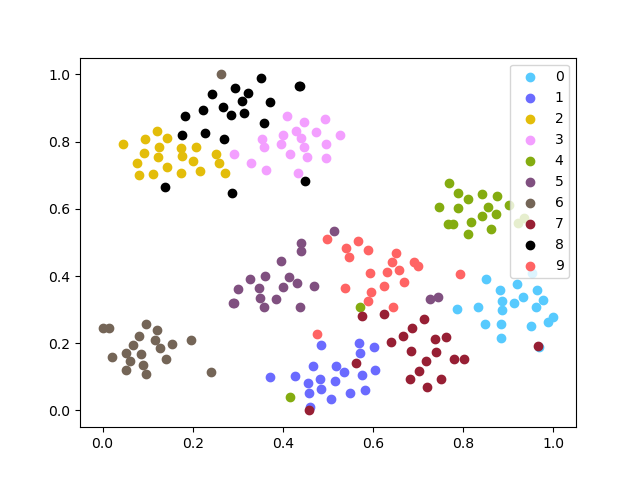}
    }
    \subfloat[SimCLR+FT]{
        \label{fig:tSNE-simclr-FT}
        \includegraphics[width=0.275\linewidth]{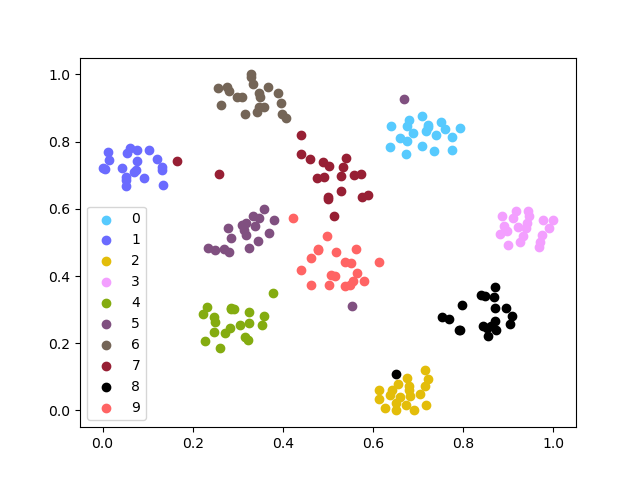}
    }
    \subfloat[IDMM+FT (ous)]{
        \label{fig:tSNE-InsDis-FT}
        \includegraphics[width=0.275\linewidth]{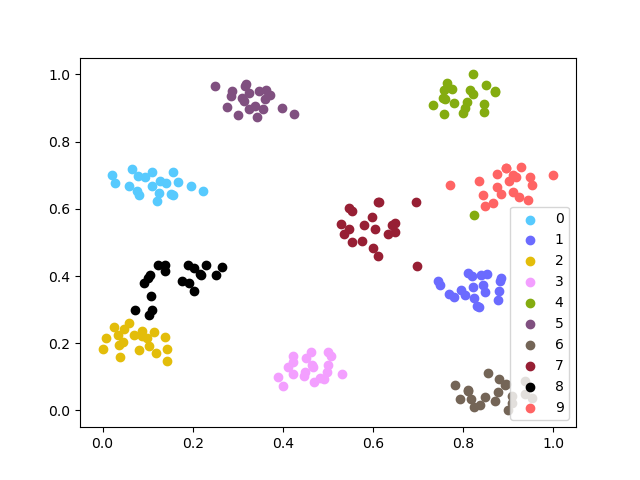}
    }
	\vspace{-4pt}
    \caption{t-SNE~\cite{tsNE} visualization of 10 classes selected from flowers using DeiT-Tiny. The first row shows the results before fine-tuning (i.e., without using any class labels) and the second row shows the results after fine-tuning (`FT'). This figure is best viewed in color.}
    \label{fig:tSNE}
	\vspace{-8pt}
\end{figure*}

Hence, when the weights $W$ are appropriately set, instance discrimination is equivalent to the deep clustering loss, which can observe potential instance similarities. As can be seen from Figure~\ref{fig:tSNE}, instance discrimination learns more distributed representations and captures better intra-class similarities when compared to other methods. 

Since in this paper we focus on ViTs, there is another important reason why we choose parametric instance discrimination: the simplicity and stability. As noted in~\cite{mocov3:chen:ICCV21}, instability is a major issue that impacts self-supervised ViT training. Hence, the form of instance discrimination (cross entropy) is more stable and easier to optimize. It will be further demonstrated in Sec.~\ref{sec:transfer} and Sec.~\ref{sec:ablation} that our method can better adapt to various emerging ViT networks and does not rely on specific designs (e.g., projection MLP head).

\subsection{Gradient Analysis}
\label{sec:gradient}

Consider the loss term for the $i$-th instance in Eq.~\eqref{eq:loss_insdis}:
\begin{equation}
    L_{\text{InsDis}}^{(i)} = -\mathbf{w}_i^T\mathbf{z}_i+\log\sum\nolimits_{j=1}^N{e^{\mathbf{w}_j^T{\mathbf{z}_i}}} \,.
\end{equation}
Then, the gradient w.r.t. $\mathbf{z}_i$ can be calculated as follows:
\begin{align}
    \frac{\partial{L}}{\partial{\mathbf{z}_i}}=&-\mathbf{w}_i+\sum_{j=1}^N\frac{e^{\mathbf{w}_j^T\mathbf{z}_i}}{\sum_{k=1}^Ne^{\mathbf{w}_k^T{\mathbf{z}_i}}}\mathbf{w}_j \\
    =&-\mathbf{w}_i+\sum\nolimits_{j=1}^N P_j^{(i)}\mathbf{w}_j\,,
\end{align}
and similarly the gradient w.r.t. $\mathbf{w}_k$ is:
\begin{equation}
\begin{aligned}
    \frac{\partial{L}}{\partial{\mathbf{w}_k}}=(P_k^{(i)}-\delta_{\{k=i\}})\mathbf{z}_i,
\end{aligned}
\end{equation}
where $\delta$ is an indicator function, equals 1 iff $k=i$.

Notice that for instance discrimination the number of classes $N$ can easily go very large and there exists extremely infrequent visiting of instance samples~\cite{ParametricInstance:cao:arxiv2020, onemillion:liu:AAAI21}. Hence for infrequent instances $k\neq{i}$, we can expect $P^{(i)}_k\approx0$ and hence $\frac{\partial{L}}{\partial{\mathbf{w}_k}} \approx \mathbf{0}$, which means extremely infrequent update of $\mathbf{w}_k$. \cite{ParametricInstance:cao:arxiv2020} and \cite{onemillion:liu:AAAI21} introduced different strategies to alleviate the problems for large datasets, such as the high GPU computation and memory overhead. Since in this paper we focus on small datasets, such strategies are not necessary. Instead, we use CutMix~\cite{cutmix:yun:ICCV19} and label smoothing~\cite{labelsmoothing:Szegedy:CVPR16} to update the weight matrix more frequently by directly modifying the one-hot label, which are also commonly used in supervised training of ViTs. If we use label smoothing, then 
\begin{equation}
\label{LS_yk}
% T^k =
y^{(i)}_c = \left\{
\begin{array}{rcl}
1-\epsilon&  & \text{if}\quad c=i,\\
\frac{\epsilon}{N-1} && \text{otherwise}
\end{array} \right.,
\end{equation}
where $\epsilon$ is the smoothing factor and we set it to 0.1 throughout this paper. Then the loss becomes:
\begin{align}
    L^{(i)}_{\text{InsDis}} =& -(1-\epsilon)\mathbf{w}_i^T\mathbf{z}_i+(1-\epsilon)\log\sum\nolimits_{j=1}^N{e^{\mathbf{w}_j^T{\mathbf{z}_i}}}\notag\\
    &-\frac{\epsilon}{N-1}\sum\nolimits_{k\neq{i}}\mathbf{w}_k^T\mathbf{z}_i+\epsilon\log\sum\nolimits_{j=1}^N{e^{\mathbf{w}_j^T{\mathbf{z}_i}}}\notag\\
    =&-(1-\epsilon)\mathbf{w}_i^T\mathbf{z}_i-\frac{\epsilon}{N-1}\sum\nolimits_{k\neq{i}}\mathbf{w}_k^T\mathbf{z}_i \notag\\
    &+\log\sum\nolimits_{j=1}^N{e^{\mathbf{w}_j^T{\mathbf{z}_i}}}
	\label{eq:labelsmoothing}
\end{align}

If we continue to use CutMix, Eq.~\eqref{eq:labelsmoothing} becomes:
\begin{align}
    L_{\text{InsDis}}^{(i)} =& -C_i\mathbf{w}_i^T\Tilde{\mathbf{z}}_{ii'}-C_{i'}\mathbf{w}_{i'}^T\Tilde{\mathbf{z}}_{ii'} \notag\\
    &-C\sum\nolimits_{j\neq{i,i'}}\mathbf{w}_j^T\Tilde{\mathbf{z}}_{ii'}+\log\sum\nolimits_{j=1}^N{e^{\mathbf{w}_j^T{\Tilde{\mathbf{z}}_{ii'}}}}\,,\notag
\end{align}
where $\lambda$ is the mixed coefficient, $i'$ is the index of the other instance in CutMix, $\Tilde{\mathbf{z}}_{ii'}$ is the output of the mixed input and
\begin{equation}
    \left\{
    \begin{array}{l}
    C_i=\lambda(1-\epsilon)+(1-\lambda)\frac{\epsilon}{N-1}\\
    C_{i'}=(1-\lambda)(1-\epsilon)+\lambda\frac{\epsilon}{N-1}\\
    C = \lambda\frac{\epsilon}{N-1}
    \end{array} \right..
\end{equation}
And the gradient w.r.t. $\mathbf{w}_k$ becomes:
\begin{align}
    \frac{\partial{L}}{\partial{\mathbf{w}_k}}=&\Big(P_k^{(ii')}-C_i\delta_{\{k=i\}}-C_{i'}\delta_{\{k=i'\}}\notag\\
    &-C(1-\delta_{\{k=i\}}-\delta_{\{k=i'\}})\Big)\Tilde{\mathbf{z}}_{ii'}\,.
\end{align}

If we set $\lambda=0.5$ and $N=2040$, then $C_i=C_{i'}\approx0.45$ and $C\approx2.5e-5$. Hence, we are able to update $\mathbf{w}_k$ even for instances $k\neq{i}$ (with relative large gradients for $\mathbf{w}_i$ and $\mathbf{w}_{i'}$ and small gradients for others), which alleviates the infrequent updating problem. Moreover, we can also alleviate the overfitting problem by using CutMix as our regularization with limited data, as revealed in~\cite{mixup:ICLR18, cutmix:yun:ICCV19}.

In conclusion, we use the following strategies to enhance instance discrimination on small datasets:

(1) Small resolution. It has been shown in~\cite{S3L:cao:arxiv2021} that small resolution during pretraining is useful for small datasets.

(2) Multi-crop. As analyzed before, instance discrimination generalizes the contrastive loss to capture both feature alignment and uniformity when using multiple crops.

(3) CutMix and label smoothing. As analyzed above, it helps us alleviate the overfitting and infrequent accessing problem when applying instance discrimination.

We name our method as \textbf{i}nstance \textbf{d}iscrimination with \textbf{m}ulti-crop and Cut\textbf{M}ix (IDMM) and we conduct ablation studies on these strategies in Sec.~\ref{sec:ablation}.

\begin{table}
	\caption{Statistics of the 6 small datasets used in the paper.}	\label{tab:dataset-overview}
	\centering
	\vspace{-6pt}
	\renewcommand{\arraystretch}{0.8}
	\footnotesize
	\renewcommand{\multirowsetup}{\centering}
	\begin{tabular}{l|c|c|c}
			\hline
			Datasets & \# Category & \# Training & \# Testing \\
			\hline 
			Flowers (\cite{flowers}) &102& 2040&6149\\
			Pets (\cite{pets}) &\pt37& 3680&3669\\
			DTD (\cite{dtd}) &\pt47& 3760&1880\\
			Indoor67 (\cite{indoor67:CVPR09}) &\pt67& 5360&1340\\
			CUB200 (\cite{cub200}) & 200 & 5994 & 5794  \\
			Aircrafts (\cite{aircrafts}) &100 & 6667&3333\\
			Cars (\cite{cars}) & 196 & 8144 & 8041 \\
			\hline
	\end{tabular}
\end{table} 

\section{Experiments}

We used 7 small datasets for our experiments, as shown in Table~\ref{tab:dataset-overview}. First, we explain the reasons why do we need training from scratch in Sec.~\ref{sec:why} and training from scratch results in Sec.~\ref{sec:scratch-results}. Then, we study the transferring ability of ViTs pretrained on small datasets (even facilitate large-scale datasets training) in Sec.~\ref{sec:transfer}. Finally, we conduct ablation studies on different components in Sec.~\ref{sec:ablation}. All our experiments were conducted using PyTorch, and we used Titan Xp GPUs for ImageNet experiments and Tesla K80 for small datasets. Codes will be made publicly available.

\begin{figure}
	\centering
	\includegraphics[width=0.75\columnwidth]{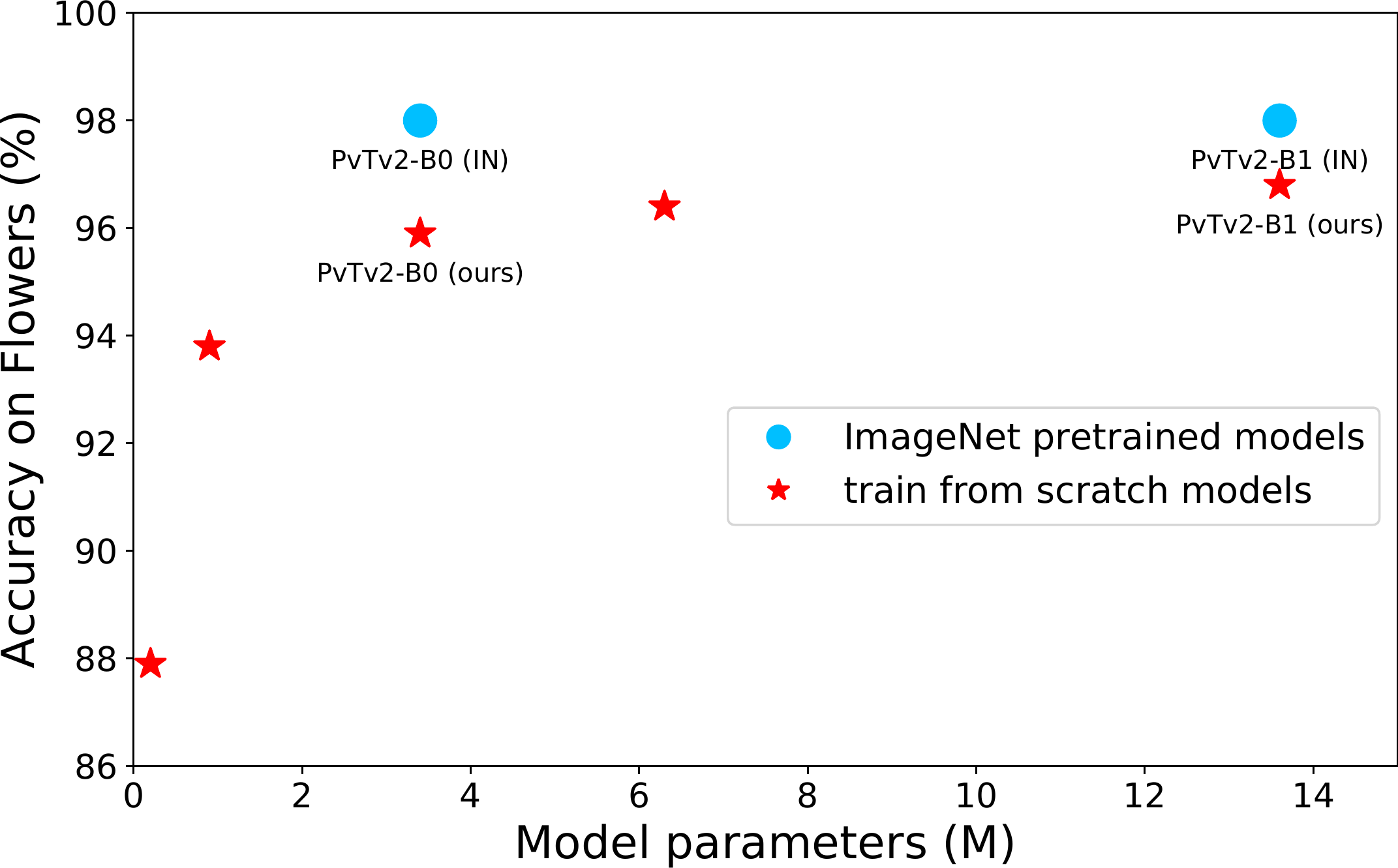}
	\caption{Parameter-Accuracy trade-off on flowers. The blue circles represent IN pretrained models while the red stars represent models of different sizes training from scratch using our method. }
	\vspace{-6pt}
	\label{fig:scatter}
\end{figure}

\begin{table*}[!htbp]
	\caption{Comparison between different pretraining methods. All fine-tuned for 200 epochs. }
	\label{tab:ssl-comparison}
	\vspace{-6pt}
	\centering
	\footnotesize
	\renewcommand{\arraystretch}{0.85}
	\renewcommand{\multirowsetup}{\centering}
	\begin{tabular}{l|l|c|c|c|c|c|c|c|c}
		\hline
		\multirow{2}{*}{Backbone} & \multicolumn{2}{c|}{pretraining} & \multicolumn{7}{c}{Accuracy} \\
		\cline{2-10}
		& method & epochs & Flowers & Pets & Dtd & Indoor67 & CUB & Aircraft & Cars\\
		\hline
		\multirow{7}{*}{DeiT-Tiny~\cite{DeiT:Touvron:ICML2021}} & random init. & 0 & 58.1&31.8 &49.4&31.0&23.8&14.6&12.3\\
		\cline{2-10}
		& SimCLR~\cite{simclr:hinton:ICML20}&\multirow{6}{*}{800}&71.1&52.1&55.9&50.7&36.2&43.2&64.3\\
		& SupCon~\cite{supcon:khosla:nips20}& &72.3&50.3&55.6&49.3&37.8&29.4&66.2\\
		& MoCov2~\cite{mocov2:xinlei:arxiv2020}& &61.8&41.5&50.6&41.1&31.6&37.7&44.0\\
		& MoCov3~\cite{mocov3:chen:ICCV21}& &67.0&52.9&52.9&49.4&20.5&32.0&53.7\\
		& DINO~\cite{dino:caron:iccv2021}&   &64.1 &51.3&51.7&46.9&41.8&\textbf{45.7}&65.3\\
		& IDMM (Ours)& &\textbf{79.9}&\textbf{56.7}&\textbf{61.2}&\textbf{53.9}&\textbf{43.1}&43.2&\textbf{66.4}\\
		\hline
	\end{tabular}
\end{table*}

\begin{table*}[!htbp]
	\caption{Training from scratch results. Both the pretraining and fine-tuning are only performed on the target dataset.}
	\label{tab:train-from-scratch}
	\vspace{-6pt}
	\centering
	\footnotesize
	\setlength{\tabcolsep}{4pt}
	\renewcommand{\arraystretch}{0.85}
	\renewcommand{\multirowsetup}{\centering}
	\begin{tabular}{l|l|c|c|c|c|c|c|c|c|c}
		\hline
		\multirow{2}{*}{Backbone} &\multirow{2}{*}{Method}& \multicolumn{2}{c|}{Fine-tuning} & \multicolumn{7}{c}{Accuracy} \\
		\cline{3-11}
		& & resolution & epochs & Flowers & Pets & Dtd & Indoor67 & CUB & Aircraft & Cars\\
		\hline
		\multirow{4}{*}{DeiT-Tiny~\cite{DeiT:Touvron:ICML2021}}& \textcolor{gray}{IN super.}&\textcolor{gray}{224} & \textcolor{gray}{200} &\textcolor{gray}{97.3}  & \textcolor{gray}{88.6}&\textcolor{gray}{73.2}&\textcolor{gray}{75.6}&\textcolor{gray}{76.8}&\textcolor{gray}{78.7}& \textcolor{gray}{90.3} \\
		\cline{2-11}
		& random init.&224 & 800 &67.8 &44.5 &54.5&40.6&24.3&33.2&38.8\\
		\cline{2-11}
		&\multirow{2}{*}{IDMM (ours)}&224&800& 83.4& 59.0& 61.8&56.1 & 45.0&46.0 &73.7\\
		& &224$\rightarrow$448&800$\rightarrow$100&\textbf{85.6}&\textbf{64.2}& \textbf{64.9}&\textbf{59.9} &\textbf{50.9}&\textbf{48.6}&\textbf{77.8}\\
		\hline
		\multirow{4}{*}{DeiT-Base~\cite{DeiT:Touvron:ICML2021}} & \textcolor{gray}{IN super.}&\textcolor{gray}{224} & \textcolor{gray}{200} &\textcolor{gray}{97.7}  & \textcolor{gray}{91.4}&\textcolor{gray}{74.9}&\textcolor{gray}{78.1}&\textcolor{gray}{81.9}&\textcolor{gray}{82.8}& \textcolor{gray}{92.6} \\
		\cline{2-11}
		 & random init. &224& 800 & 67.3 & 48.4&46.0&44.0&27.7  &30.1&33.3\\
		 \cline{2-11}
		&\multirow{2}{*}{IDMM (ours)}&224&800& 88.1& 63.2&62.3 & 57.4&47.8 &43.1 &64.5\\
		& &224$\rightarrow$448&800$\rightarrow$100&\textbf{90.6}&\textbf{67.2}& \textbf{67.3}& \textbf{61.7}&\textbf{54.3}&\textbf{46.6}&\textbf{70.7}\\
		\hline
		\multirow{4}{*}{PVTv2-B0~\cite{pvtv2:wang:arxiv2021}}  & \textcolor{gray}{IN super.}&\textcolor{gray}{224} & \textcolor{gray}{200} &\textcolor{gray}{98.0}  & \textcolor{gray}{90.5}&\textcolor{gray}{75.9}&\textcolor{gray}{76.7}&\textcolor{gray}{81.4}&\textcolor{gray}{88.3}& \textcolor{gray}{92.6} \\
		\cline{2-11}
		 & random init. &224& 800 &  90.3&80.5 &57.7&66.3&66.6  &74.8&87.9\\
		\cline{2-11}
		&\multirow{2}{*}{IDMM (ours)}&224&800&94.6&84.7&69.3&69.6&73.8&79.8&90.9\\
		& &224$\rightarrow$448&800$\rightarrow$100&\textbf{95.9} &\textbf{88.0} & \textbf{73.2}&\textbf{73.7} &\textbf{77.6}&\textbf{83.3}&\textbf{92.0}\\
		\hline
		\multirow{4}{*}{PVTv2-B3~\cite{pvtv2:wang:arxiv2021}} & \textcolor{gray}{IN super.}&\textcolor{gray}{224} & \textcolor{gray}{200} &\textcolor{gray}{98.7}  & \textcolor{gray}{93.6}&\textcolor{gray}{78.1}&\textcolor{gray}{80.8}&\textcolor{gray}{85.5}&\textcolor{gray}{91.7}& \textcolor{gray}{94.4} \\
		\cline{2-11}
		 & random init. &224& 800 & 90.5  &83.4  & 64.5& 67.5&66.2   &85.0&89.9\\
		\cline{2-11}
		&\multirow{2}{*}{Ours}&224&800& 95.9 &89.8 &68.9&73.2& 79.0 &90.5&94.0\\
		& &224$\rightarrow$448&800$\rightarrow$100&\textbf{96.7} &\textbf{91.9} & \textbf{71.8}&\textbf{76.3}  &\textbf{82.8} &\textbf{91.8} & \textbf{94.3}\\
		\hline
		\multirow{4}{*}{T2T-ViT-7~\cite{t2t:yuan:arxiv2021}} & \textcolor{gray}{IN super.}&\textcolor{gray}{224} & \textcolor{gray}{200} &\textcolor{gray}{97.7}  & \textcolor{gray}{90.5}&\textcolor{gray}{75.2}&\textcolor{gray}{76.6}&\textcolor{gray}{79.0}&\textcolor{gray}{83.8}& \textcolor{gray}{92.8} \\
		\cline{2-11}
		 & random init. &224& 800 & 82.1 &66.2&58.5 &57.7 &35.7  & 57.2&60.3 \\
		\cline{2-11}
		&\multirow{2}{*}{IDMM (ours)}&224&800& 90.8  &  75.0&64.7 &66.0 &59.0 &71.4& 89.9\\
		& &224$\rightarrow$448&800$\rightarrow$100&\textbf{91.7} &\textbf{76.9} & \textbf{65.7}&\textbf{68.9}  &\textbf{63.2} &\textbf{72.9} & \textbf{91.2}\\
		\hline
	\end{tabular}
\end{table*}

\subsection{Why training from scratch?}
\label{sec:why}

We explain the reasons why do we need training from scratch directly on target datasets from 3 aspects:
\squishlist
	\item \textbf{Data.} Current ViT models are often pretrained on a large-scale dataset (such as ImageNet or even larger ones), and then fine-tuned in various downstream tasks. Moreover, the lack of the typical convolutional inductive bias makes these models more data-hungry than common CNNs. Hence, it is critical to investigate whether we can train ViTs from scratch for a task where the \emph{total} amount of available images is limited (e.g., 100 categories with roughly 20 images per category).
	\item \textbf{Computing.} The combination of a large-scale dataset, a large number of epochs and a complex backbone network means that ViT training are extremely computationally expensive. This phenomenon makes ViT a privilege for researchers at few institutions.
	\item \textbf{Flexibility.} The pretraining followed by downstream fine-tuning paradigm will sometimes become cumbersome. For instance, we may need to train 10 different models for the same task, and deploy them to different hardware platforms~\cite{once-for-all:hansong:ICLR20}, but it is impractical to pretrain 10 models on a large-scale dataset.
\squishend

As shown in Fig.~\ref{fig:scatter}, it is obvious that ImageNet pretrained models need much more data and computational cost when compared to training from scratch. Moreover, when we need to deploy models of different sizes on terminal devices, training from scratch provides better parameter-accuracy trade-offs. For instance, the smallest ImageNet pretrained model of PVTv2 (i.e., B0) has 3.4M parameters, which may still be too big for some devices. In contrast, we can train a much smaller model (0.8M) from scratch to adapt to the devices, which reaches 93.8\% accuracy using our IDMM.

\subsection{Training from scratch results}
\label{sec:scratch-results}

In this section, we investigate training ViTs from scratch. Following~\cite{S3L:cao:arxiv2021}, the full learning process contains two stages: pretraining and fine-tuning. We use the pretrained weights obtained by SSL for initialization and then  fine-tune networks for classification using the cross entropy loss. Note that \emph{SSL pretraining and fine-tuning are both performed \emph{only} on the target dataset}. Our method focuses on the first stage and the fine-tuning stage follows common practices.

For the fine-tuning stage, we follows the setup in DeiT~\cite{DeiT:Touvron:ICML2021} and fine-tune all methods for 200 epochs (except for Table~\ref{tab:train-from-scratch}). Specifically, we use AdamW with a batch size of 256 and a weight decay of 1e-3. The learning rate (lr) is initialized to 5e-4 and follows the cosine learning rate decay. For the SSL pretraining stage, all methods are pretrained for 800 epochs and our IDMM follows the same training settings as in the fine-tuning stage. We set $\alpha=0.5$ for CutMix in our IDMM. We follow the settings in the original papers for other methods and more details are included in the appendix. We use 112x112 resolution during pretraining and 224x224 during fine-tuning for all methods, as suggested in~\cite{S3L:cao:arxiv2021}.

\begin{figure}[!htbp]
	\centering
	\includegraphics[width=0.75\columnwidth]{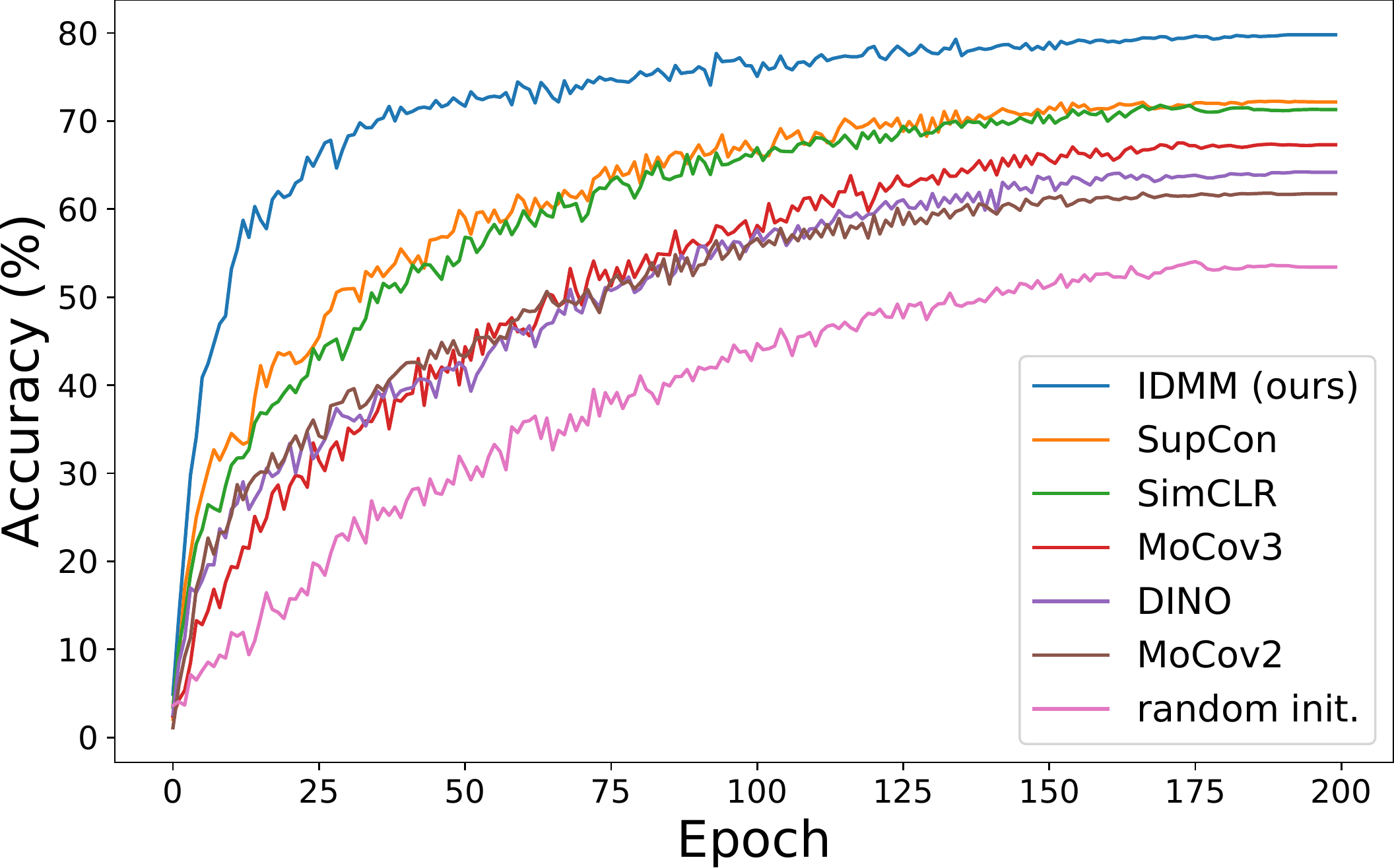}
	\vspace{-6pt}
	\caption{Comparison of different SSL methods on flowers dataset. All pretrained and fine-tuned ony on flowers for the same epochs.}
	\label{fig:comparison}
\end{figure}

\begin{table}[!htbp]
	\caption{Standard deviation of our method. We report the mean and standard deviation of 3 runs for both stages.}
	\label{tab:std}
	\centering
	\vspace{-6pt}
	\footnotesize
	\renewcommand{\arraystretch}{0.9}
	\renewcommand{\multirowsetup}{\centering}
	\begin{tabular}{l|c|c|c}
		\hline
		Backbone & Stage& Flowers&Pets \\
		\hline
		\multirow{2}{*}{PVTv2-B0} &pretraining&92.5$\pm$0.1&83.1$\pm$0.3\\
		\cline{2-4}
		& fine-tuning&92.4$\pm$0.2&83.5$\pm$0.2\\
		\hline
		\multirow{2}{*}{T2T-ViT-7} &pretraining&89.0$\pm $0.3&70.9$\pm$0.3\\
		\cline{2-4}
		& fine-tuning&88.6$\pm $0.1&70.3$\pm$0.2\\
		\hline
	\end{tabular}
\end{table}

\begin{table*}[!htbp]
	\caption{Transferring ability when pretrained on small datasets. The element with the highest accuracy in each cell and column is \underline{underlined} and \textbf{bolded}, respectively}
	\label{tab:transfer}
	\centering
	\vspace{-6pt}
	\footnotesize
	\renewcommand{\arraystretch}{0.9}
	\renewcommand{\multirowsetup}{\centering}
	\begin{tabular}{l|c|c|c|c|c|c|c|c|c}
		\hline
		\multirow{2}{*}{Backbone} & \multicolumn{2}{c|}{Pretraining} & \multicolumn{7}{c}{Transferring Accuracy} \\
		\cline{2-10}
		& Datasets & Method & Flowers & Pets & Dtd&Indoor67  & CUB &Aircraft& Cars\\
		\hline
		\multirow{22}{*}{PVTv2-B0} & \multirow{3}{*}{Flowers} &IDMM &\cellcolor{lightgray}{\underline{92.4}} &\underline{83.1}& \underline{64.8}&\underline{66.3}&\underline{69.9}&\underline{77.1}&\underline{87.3}\\
		&&SimCLR &\cellcolor{lightgray}{90.1} &80.7 &61.6&64.3&62.3&72.8&86.6\\
		&&SupCon &\cellcolor{lightgray}{91.2} &82.4 &63.1&65.3&66.3&75.0&87.0\\
		\cline{2-10}
	     & \multirow{3}{*}{Pets} &IDMM &\underline{92.8} &\cellcolor{lightgray}{83.2} &\underline{65.3}&\underline{64.9}&\underline{70.1}&\underline{78.1}&87.3\\
		&&SimCLR &89.9 &\cellcolor{lightgray}{82.8} &62.7&63.7&67.6&76.1&86.6\\
		&&SupCon &90.4 &\cellcolor{lightgray}{\underline{84.7}} &63.5&64.6&69.6&76.1&\underline{87.8}\\
        \cline{2-10}
        & \multirow{3}{*}{Dtd} &IDMM & \underline{92.9}&\underline{82.9} &\cellcolor{lightgray}{\underline{66.9}}&\underline{67.3}&\underline{70.0}&\underline{78.5}&\underline{86.7}\\
		&&SimCLR &89.1 & 79.4&\cellcolor{lightgray}{62.3}&64.0&64.4 &73.9&85.4\\
		&&SupCon & 88.9 & 79.7&\cellcolor{lightgray}{62.3}&63.6& 65.1&75.8&86.2\\
		\cline{2-10}
	     & \multirow{3}{*}{Indoor67} &IDMM & \underline{93.2} &  \underline{82.7}&\underline{65.4} &\cellcolor{lightgray}{\underline{68.5}}& \underline{70.4} &\textbf{\underline{79.7}}& \underline{87.7}\\
		&&SimCLR & 90.3 & 80.7 &62.8 &\cellcolor{lightgray}{66.6}& 61.3 &72.8& 86.4\\
		&&SupCon &90.9  & 82.2 &62.9 &\cellcolor{lightgray}{65.0}& 66.9 &74.6& 86.8\\
        \cline{2-10}
	     & \multirow{3}{*}{CUB} &IDMM &\underline{\textbf{93.7}}  & \underline{83.3}&\textbf{\underline{67.0}}&\underline{68.7}& \cellcolor{lightgray}{\underline{69.8}}&\underline{78.7}&\underline{87.6}\\
		&&SimCLR &91.3& 82.2&63.9&64.9&\cellcolor{lightgray}{68.5}& 76.7&87.3\\
		&&SupCon &90.6& 83.0&63.8&66.5&\cellcolor{lightgray}{68.6}&77.0 &87.4\\
		\cline{2-10}
	     & \multirow{3}{*}{Aircraft} &IDMM &\underline{91.3}  & \underline{82.0} & \underline{64.5}&\underline{64.3}& \underline{70.3} &\cellcolor{lightgray}{73.4}& \underline{87.3}\\
		&&SimCLR & 87.0 & 78.3 &60.6 &62.9&65.2 &\cellcolor{lightgray}{74.4}&86.2 \\
		&&SupCon & 87.9 &79.3  & 62.4&61.9& 66.4 &\cellcolor{lightgray}{\underline{76.5}}& 86.2\\
		\cline{2-10}
	     & \multirow{3}{*}{Cars} &IDMM & \underline{93.4} &\textbf{\underline{85.0}} &\underline{66.5}&\textbf{\underline{69.4}}&\textbf{\underline{72.2}} &\underline{79.5}&\cellcolor{lightgray}{87.8}\\
		&&SimCLR &90.9& 84.5&64.3&67.4&68.8&79.1&\cellcolor{lightgray}{89.3}\\
		&&SupCon &91.1&84.6 &65.1&68.3&70.4&79.3&\cellcolor{lightgray}{\textbf{\underline{90.6}}}\\
		\cline{2-10}
        &N/A&random init.&76.3&65.1&55.7&58.9&55.2&41.7&76.7\\
		\hline
	\end{tabular}
\end{table*}

First, we compare our method with popular SSL methods for both CNNs and ViTs in Table~\ref{tab:ssl-comparison}. For fair comparisons, all methods are pretrained for 800 epochs and then fine-tuned for 200 epochs. As can be seen in Table~\ref{tab:ssl-comparison} and Figure~\ref{fig:comparison}, SSL pretraining is useful even when training from scratch and all SSL methods perform better than random initialization. Our method achieves the 
highest accuracy on all these datasets, except for aircraft. When the number of images is small (e.g., flowers and pets), the advantage of our method is more obvious, which is consistent to our analyses before.

Then, following~\cite{S3L:cao:arxiv2021}, we fine-tune the models for longer epochs to get better results. Specifically, with the IDMM initialized weights, we first fine-tune for 800 epochs under 224x224 resolution and then continue fine-tuning for 100 epochs under 448x448 resolution. As shown in Table~\ref{tab:train-from-scratch}, we achieve \textbf{the state-of-the-art results when training from scratch} on these 7 datasets for all these ViT models, to the best of our knowledge. Moreover, the gap between training from scratch and using ImageNet pretrained models (colored in gray) has been greatly reduced using our method, which indicates that training from scratch is promising even for ViT models. Notice that PVTv2 models achieve better performance than DeiT and T2T by introducing convolutions to ViTs. The introduction of the typical convolutional inductive bias makes it less data-hungry than common ViTs and hence achieving better performance on these small datasets.

Further, we also investigate the randomness during both the pretraining and fine-tuning stage because the number of training images is small. For the pretraining stage, we pretrain 3 different models (using our method) and fine-tune these models separately. For the fine-tuning stage, we fine-tune 3 times with one pre-trained model. As shown in Table~\ref{tab:std}, the standard deviation is small in both stages on the two smallest datasets and hence we only report single run results in Table~\ref{tab:ssl-comparison} and \ref{tab:train-from-scratch}.

\begin{table*}[!htbp]
	\caption{Transferring ability when pretrained on 10,000 images from ImageNet. All elements are obtained by finetuning for 200 epochs.}
	\label{tab:sin_transfer}
	\centering
	\vspace{-6pt}
	\footnotesize
	\renewcommand{\arraystretch}{0.9}
	\renewcommand{\multirowsetup}{\centering}
	\begin{tabular}{l|c|c|c|c|c|c|c|c|c}
		\hline
		\multirow{2}{*}{Backbone} & \multicolumn{2}{c|}{Pretraining} & \multicolumn{6}{c}{Transferring Accuracy} \\
		\cline{2-10}
		& Datasets & Method & Flowers & Pets & Dtd &Indoor67 & CUB &aircraft& Cars\\
		%\hline
		%PvTv2-B0 & small IN-10k (gray edge) & InsDis & 91.4 & 82.6  & 64.5  &65.1  & 66.3& 78.7&88.1 \\
		\hline
		\multirow{4}{*}{PVTv2-B0} & \multirow{4}{*}{SIN-10k} &IDMM &\textbf{93.8}&\textbf{83.6} &\textbf{66.8}& \textbf{69.4}&\textbf{70.7} &\textbf{81.3}&\textbf{87.5}\\
		\cline{3-10}
		&&MoCov3 &91.0&81.4&62.3&66.3&63.7&74.5&86.2\\
		\cline{3-10}
		&&DINO &92.3&82.3&65.9&68.5&65.8&76.9&86.4\\
		\cline{3-10}
		&&supervised&92.9&81.7&66.1&65.9&66.6&78.7&86.0\\
		\hline
		%\hline
		%PvTv2-B3 & small IN-10k (gray) & InsDis &95.8 & 87.7 & 68.9 &72.2 &73.1&89.9&93.6\\
    	%\hline
		\multirow{4}{*}{PVTv2-B3} & \multirow{3}{*}{SIN-10k} &IDMM &\textbf{95.9} &  \textbf{88.4}& \textbf{70.1} & \textbf{73.6}& \textbf{76.8}&\textbf{87.5}&\textbf{92.9} \\
		\cline{3-10}
		&&MoCov3& 93.7&87.1& 66.0& 70.5&63.7&82.2 &92.3\\
		\cline{3-10}
		&&DINO &95.0&87.8&68.3&73.4&72.4&86.1&92.5\\
		\cline{3-10}
		&&supervised&90.9&80.9&62.9 & 63.3&65.6&83.8 &89.7\\
	    \hline
	    \multirow{2}{*}{T2T-ViT-7} & \multirow{2}{*}{SIN-10k} &IDMM &  \textbf{89.8}&\textbf{74.1}&\textbf{63.5}&\textbf{62.6}&\textbf{55.2}&\textbf{72.7}&\textbf{82.4} \\
		\cline{3-10}
		&&supervised& 80.8& 57.8&57.5 & 50.7&35.6&56.8 &59.9\\
	    \hline
	\end{tabular}
\end{table*}

\subsection{Transfer ability of small datasets}
\label{sec:transfer}

Having investigated training from scratch on small datasets for ViT models, we now study the transfer ability of the representations learned on these small datasets. The transfer ability of representations pretrained on large-scale datasets has been well studied, but few works studied the transfer ability of small datasets.

In Table~\ref{tab:transfer} we evaluate the transferring accuracy of models pretrained on different datasets. As in Sec.~\ref{sec:scratch-results}, we train 800 epochs for pretraining and fine-tuning 200 epochs. The on-diagonal cells (colored gray) perform pretraining and fine-tuning on the same dataset. The off-diagonal cells evaluate transfer performance across these small datasets. From Table~\ref{tab:transfer} we can have the following observations:
\squishlist
    \item ViTs have good transferring ability even when pretrained on small datasets. This means that we can use pretrained models from small datasets to transfer to other datasets in different domains to improve performance. 
    
    \item Our method also has higher transferring accuracy on all these datasets when compared to SimCLR and SupCon. As analyzed before, we think that it is due to the learnable fully connected layer $W$, which can capture both feature alignment and instance similarity. Also, the learnable fc better protects features from learning specific properties of the loss, as will be shown in Sec.~\ref{sec:ablation}.
    
    \item We can obtain surprisingly good results even if the pretrained dataset and the target dataset are \emph{not} in the same domain. For instance, models pretrained on Indoor67 achieve the highest accuracy when transfer to Aircraft. It is obvious that the number of images in the pretrained dataset matters, because Cars performs best in all. However, we want to argue that it is not the only reason because we can see that Indoor67 and CUB perform better than Cars in some cases despite having fewer training images. We leave it to future work to study what properties matter for pretraining datasets when transferring.
\squishend

\begin{table}[!htbp]
	\caption{Top-1 accuracy (\%) on ImageNet.}
	\label{tab:imagenet}
	\centering
	\vspace{-6pt}
	\footnotesize
	\setlength{\tabcolsep}{2.5pt}
	\renewcommand{\arraystretch}{0.9}
	\renewcommand{\multirowsetup}{\centering}
	\begin{tabular}{l|c|c|c}
		\hline
		Backbone & Method & Epochs& Acc. (\%) \\
		\hline
		\multirow{6}{*}{PVTv2-B0} &random init. &\multirow{4}{*}{100}&68.6\\
		& MoCov3 (SIN-10k) &  & 68.8\\
		& IDMM (SIN-10k) &  & \textbf{69.5}\\
		& IDMM (SIN-total 10k) &  & \textbf{69.5}\\
		\cline{2-4}
		& random init. & \multirow{2}{*}{300} & 70.0\\
		& IDMM (SIN-10k) &  & \textbf{70.9}\\
        \hline
		\multirow{4}{*}{DeiT-Tiny} &random init. &\multirow{2}{*}{100}&66.8\\
		& IDMM (SIN-10k) &  & \textbf{67.8}\\
		\cline{2-4}
		& random init. & \multirow{2}{*}{300} & 72.2\\
		& IDMM (SIN-10k) &  & \textbf{72.9}\\
		\hline
	\end{tabular}
\end{table}

After observing that models pretrained on small datasets have surprisingly good transferring ability, we can further explore the potential of small datasets. We sample the original ImageNet to smaller subsets with 10,000 images (SIN-10k), motivated by~\cite{S3L:cao:arxiv2021}. By pretraining models on SIN-10k, we evaluate the performance when transferring to small datasets in Table~\ref{tab:sin_transfer} as well as the large-scale dataset ImageNet in Table~\ref{tab:imagenet}. In Table~\ref{tab:sin_transfer} we compare our method with various SSL methods as well as the supervised baseline under different backbones. It can be seen that our method has a large edge over these comparison methods and representations learned on SIN-10k can serve as a good initialization when transferring to other datasets. It is worth noting that MoCov3 and DINO fail to converge under T2T-ViT-7 after trying various hyper-parameters so we don't report the results for them in Table~\ref{tab:sin_transfer}. It indicates our method can be easily applied to emerging ViTs without the need of special design or tuning. 

Furthermore, we investigate whether we can benefit from pretraining on 10,000 images when training on ImageNet. As can be seen in Table~\ref{tab:imagenet}, using the representation learned from 10,000 images as initialization can greatly accelerate the training process and finally achieve higher accuracy (about 1 point) on ImageNet. Notice that we sampled a balanced subset before (10 images per class) and we also compare with the setting where we randomly sample 10,000 images without using label information (SIN-total 10k). As can be seen, whether to use labels when sampling (balanced or not) has no effect on the result, as noted in~\cite{sslimbalaced:yang:NIPS20}.

\subsection{Ablation studies}
\label{sec:ablation}

In this section, we first investigate the effect of different components in our method in Table~\ref{tab:ablation}. Then, we investigate the effect of the projection MLP head in Table~\ref{tab:mlp}. 

As can be seen in Table~\ref{tab:ablation}, all the 4 strategies are useful and combining all these strategies achieves the best results. The experimental results further confirm the analyses in Sec.~\ref{sec:analysis} that using multiple views and CutMix is helpful.

In Table~\ref{tab:mlp}, all methods are pretrained for 800 epochs on SIN-10k and then fine-tuned for 200 epochs when transferring to target datasets. The projection MLP head is essential for contrastive methods like SimCLR while it is not the case for instance discrimination. It further confirms the analyses in Sec.~\ref{sec:analysis} that the learnable fc $W$ protects features from learning specific properties of the loss and hence achieving better transferring ability. In contrast, the $W$ in contrastive loss is not learnable and they need extra projection head.

\begin{table}[!htbp]
	\caption{Ablation studies when training from scratch on flowers.}
	\label{tab:ablation}
	\centering
	\vspace{-6pt}
	\footnotesize
	\setlength{\tabcolsep}{2.5pt}
	\renewcommand{\arraystretch}{0.9}
	\renewcommand{\multirowsetup}{\centering}
	\begin{tabular}{l|c|c|c|c|c}
		\hline
		method&LS&small res.&multi-crop & CutMix & Acc. (\%) \\
		\hline
		\multirow{5}{*}{InsDis}
		&$\times$&$\times$&$\times$&$\times$&69.6 \\
		&\checkmark&$\times$&$\times$&$\times$&70.4 \\
		&\checkmark&\checkmark&$\times$&$\times$&73.1\\
		&\checkmark&\checkmark&\checkmark&$\times$&76.9\\
		&\checkmark&\checkmark&\checkmark&\checkmark&\textbf{79.9}\\
        \hline
	\end{tabular}
	\vspace{-6pt}
\end{table}

\begin{table}[!htbp]
	\caption{The effect of MLP head. All pretrained on SIN-10k.}
	\label{tab:mlp}
	\centering
	\vspace{-6pt}
	\footnotesize
	\setlength{\tabcolsep}{2.5pt}
	\renewcommand{\multirowsetup}{\centering}
	\begin{tabular}{l|c|c|c|c|c|c}
		\hline
		backbone&method&proj. MLP&flowers & pets &  dtd&cub\\
		\hline
		\multirow{4}{*}{DeiT-Tiny}&\multirow{2}{*}{IDMM}&$\times$&\textbf{86.6}&\textbf{65.3}&\textbf{59.1}&47.6\\
		&&\checkmark&85.2&65.1&57.8&\textbf{48.0}\\
        \cline{2-7}
        &\multirow{2}{*}{SimCLR}&$\times$&82.2&60.2&57.8&41.7\\
		&&\checkmark&\textbf{83.3}&\textbf{62.1}&\textbf{58.9}&\textbf{45.8}\\
		\hline
	\end{tabular}
	\vspace{-6pt}
\end{table}

\section{Conclusions}

In this paper, we proposed a method called IDMM for (pre)training ViTs with small data and the effectiveness of the proposed approach is well validated by both theoretical analyses and experimental studies. We achieved state-of-the-art results on 7 small datasets under various ViT backbones when training from scratch. Moreover, we studied the transferring ability of small datasets and found that ViTs also have good transferring ability even when pre-trained on small datasets. However, there is still room for improvement when training from scratch on these small datasets for architectures like DeiT. Furthermore, it is still unknown what properties matter for pretraining on small datasets when transferring and we leave them to future work.

\clearpage
{\small
\bibliographystyle{ieee_fullname}
\bibliography{egbib}
}

\clearpage
\appendix

\section{Detailed Derivations of Eq. 6\&7}
From the main paper we know that
\begin{equation}
\label{eq:appendix-insdis}
    L_{\text{InsDis}}^{(i)} = -\mathbf{w}_i^T\mathbf{z}_i+\log\sum_{j=1}^Ne^{\mathbf{w}_j^T \mathbf{z}_i}
\end{equation}

Now if we use 2 views ($\mathbf{x}_{iA}$ and $\mathbf{x}_{iB}$) for each instance in instance discrimination, then from Eq.~\eqref{eq:appendix-insdis} we have:
\begin{equation}
\begin{aligned}
    L^{(i)}_{\text{InsDis}}=&-\mathbf{w}_i^T \mathbf{z}_{iA}+\log\sum_{j=1}^Ne^{\mathbf{w}_j^T \mathbf{z}_{iA}}\\
    &-\mathbf{w}_i^T \mathbf{z}_{iB}+\log\sum_{j=1}^Ne^{\mathbf{w}_j^T \mathbf{z}_{iB}}\,,
\end{aligned}
\end{equation}

where $\mathbf{z}_{iA}$ and $\mathbf{z}_{iB}$ are corresponding representations of $\mathbf{x}_{iA}$ and $\mathbf{x}_{iB}$. If we set $\mathbf{w}_j=\mathbf{z}_{jA}$ for $\forall j$, we have:

%\begin{equation}
\begin{align*}
    L^{(i)}_{\text{InsDis}}=&-\mathbf{z}_{iA}^T \mathbf{z}_{iA}+\log\sum_{j=1}^Ne^{\mathbf{z}_{jA}^T \mathbf{z}_{iA}}\\
    &-\mathbf{z}_{iA}^T \mathbf{z}_{iB}+\log\sum_{j=1}^Ne^{\mathbf{z}_{jA}^T \mathbf{z}_{iB}}\\
    =&-\mathbf{z}_{iA}^T \mathbf{z}_{iA}+\log\left(e^{\mathbf{z}_{iA}^T \mathbf{z}_{iA}}+\sum_{j\neq{i}}e^{\mathbf{z}_{iA}^T \mathbf{z}_{jA}}\right)\\
    &-\mathbf{z}_{iA}^T \mathbf{z}_{iB}+\log\left(e^{\mathbf{z}_{iA}^T \mathbf{z}_{iB}}+\sum_{j\neq{i}}e^{\mathbf{z}_{jA}^T \mathbf{z}_{iB}}\right)\\
    =&\underbrace{-\mathbf{z}_{iA}^T \mathbf{z}_{iB}}_{\text{alignment}}
    +\underbrace{\log\left(e^{\mathbf{z}_{iA}^T \mathbf{z}_{iB}}+\sum_{j\neq{i}}e^{\mathbf{z}_{iB}^T \mathbf{z}_{jA}}\right)}_{\text{uniformity}}\\
    &\underbrace{-\mathbf{z}_{iA}^T \mathbf{z}_{iA}}_{\text{constant}}+\underbrace{\log\left(e^{\mathbf{z}_{iA}^T \mathbf{z}_{iB}}+\sum_{j\neq{i}}e^{\mathbf{z}_{iA}^T \mathbf{z}_{jA}}\right)}_{\text{uniformity}}
\end{align*}
%\end{equation}

\section{Training details}
The training details for MoCov2, MoCov3, SimCLR, SupCon and DINO in Table~\ref{tab:ssl-comparison} in the paper are shown in Table~\ref{tab:appendix-details}.

\begin{table}[t]
	\caption{Training details for MoCov2, MoCov3, SimCLR, SupCon and DINO. `bs' denotes batch size, `lr' denotes learning rate, `wd' denotes weight decay, `dim' denotes the dimension of feature, $\tau$ denotes the temperature parameter and $k$ denotes the size of memory bank in MoCov2.}
	\label{tab:appendix-details}
	\centering
	\small
	\setlength{\tabcolsep}{3.5pt}
	\begin{tabular}{l|c|c|c|c|c|c|c|c}
		\hline
		\multirow{2}{*}{Method} &  \multicolumn{8}{c}{Settings}  \\
		\cline{2-9}
		&opt&bs&lr&wd&dim&schedule&$\tau$&k\\
		\hline
		MoCov2 &SGD&256&5e-4&0.001&256&cosine&0.2&2048\\ 
        \hline
		MoCov3 &adamW&256&1.5e-4&0.1&256&cosine&0.2&-\\ 
		\hline
	   SimCLR&SGD&512&5e-4&0.001&256&cosine&0.1&-\\
		\hline
		Supcon&SGD&512&5e-4&0.001&256&cosine&0.1&-\\
		\hline
		DINO&adamW&512&5e-4&0.04&256&cosine&0.1&-\\
		\hline
	\end{tabular}
\end{table}

\end{document}